\documentclass[letterpaper, 10pt, conference]{IEEEtran}  %

\IEEEoverridecommandlockouts
\usepackage{float} 
\usepackage{multirow}
\usepackage{graphicx}
\usepackage{amsmath}
\usepackage{amsthm}
\usepackage{amsfonts}
\usepackage{dsfont}
\usepackage{subfigure}
\usepackage{booktabs}
\usepackage{algorithm}
\usepackage{algorithmic}
\usepackage{textcomp}
\usepackage{xcolor}
\usepackage{cite}
 \usepackage{booktabs} 
\def\BibTeX{{\rm B\kern-.05em{\sc i\kern-.025em b}\kern-.08em
    T\kern-.1667em\lower.7ex\hbox{E}\kern-.125emX}}

\begin{document}

\title{Shaping Sparse Rewards in Reinforcement Learning: A Semi-supervised Approach}


\author{Wenyun Li$^{1,2}$, Wenjie Huang$^{2,3}$ and Chen Sun$^{2}$
	\thanks{*This work was not supported by any organization}
    \thanks{$^{1}$ Department of Mathematics, The University of Hong Kong (HKU)}
	\thanks{$^{2}$ Department of Data and Systems Engineering, HKU}
    \thanks{$^{3}$ Musketeers Foundation Institute of Data Science, HKU}
    \thanks{*Corresponding author: Wenyun Li \tt\small wenyunli@hku.hk}%
}

\maketitle

\begin{abstract}
In many real-world scenarios, reward signal for agents are exceedingly sparse, making it challenging to learn an effective reward function for reward shaping. To address this issue, the proposed approach in this paper performs reward shaping not only by utilizing non-zero-reward transitions but also by employing the \emph{Semi-Supervised Learning} (SSL) technique combined with a novel data augmentation to learn trajectory space representations from the majority of transitions, {i.e}., zero-reward transitions, thereby improving the efficacy of reward shaping.  Experimental results in Atari and robotic manipulation demonstrate that our method outperforms supervised-based approaches in reward inference, leading to higher agent scores. Notably, in more sparse-reward environments, our method achieves up to twice the peak scores compared to supervised baselines. The proposed entropy  augmentation enhances performance, showcasing a 15.8\% increase in best score over other augmentation methods.
\end{abstract}


\section{Introduction}
The sparse reward problem is a core challenge in Reinforcement Learning (RL) when solving real-world tasks \cite{Kober2013ReinforcementLI}.
Many real-world tasks naturally have the feature of delayed or infrequent rewards due to the complexity and nature of the tasks.
Such sparse reward signals provide minimal immediate feedback to guide the agent’s exploration, while also inducing high variance in returns. These properties make it challenging to learn an optimal policy effectively \cite{Plappert2018MultiGoalRL}.

In general, densifying reward with \textit{reward design} or \textit{reward shaping} \cite{Ng1999PolicyIU} methods face the inescapable predicament of reward hacking \cite{NIPS2017_32fdab65}. This phenomenon can lead to a shift in the optimal policy, or cause the agent to learn a behavior that deviates from the intended design. In response to this dilemma, supervised-based reward shaping methods demonstrate greater fidelity to the original reward formulation. Moreover, \cite{Memarian2021SelfSupervisedOR} prove that, under mild assumptions, specific modifications to the reward function can still preserve the same set of optimal policies.
However, methods based on non-expert demonstrations under supervised learning suffer from poor sample efficiency \cite{Peng2019AdvantageWeightedRS}, especially when non-zero-reward transitions are extremely rare in sparse reward cases. 

In order to bridge the gap in handling sparse rewards, our approach performs reward shaping not only by utilizing non-zero-reward transitions but also by employing \emph{Semi-Supervised Learning} (SSL) techniques to learn trajectory space representations from the majority of transitions, \textit{i.e.}, zero-reward transitions, thereby improving reward shaping. Explicitly, we propose an RL algorithm that applies the idea of SSL, called Semi-Supervised Reward Shaping framework (SSRS), which alternates between policy updates using a base RL algorithm and transition-level reward shaping via a learned reward estimator. Consistency regularization \cite{Bachman2014LearningWP} to zero-reward transitions is adopted for the optimization of the reward estimator. 

The rationale of SSRS is substantiated by identifying state-action pairs analogous to transitions with actual environmental rewards, and by conferring denser rewards accordingly upon those deemed sufficiently confident as “good” state-action pairs. Our empirical investigation reveals that the trajectory space conforms to the mild assumptions requisite for SSL, thereby permitting the application of SSL methodologies to refine the evaluative efficacy of the inferred reward function over zero-reward transitions.

The main contributions of the paper are listed as below:
\begin{itemize}
    \item We propose a novel reward shaping framework named SSRS, which operates on any base RL algorithm and employs SSL techniques to improve sample efficiency in sparse-reward environments.
    
    \item We empirically examine the underlying conditions for SSL to be effective in RL settings, propose an additional data perturbation method to better process agent observations, and suggest potential directions for future applications of SSL in reinforcement learning.
    
    \item We evaluate the performance of SSRS in reward-sparse Atari and robotic manipulation environments, comparing it with other supervised-based approaches—SORS \cite{Memarian2021SelfSupervisedOR}, RCP \cite{Kumar2019RewardConditionedP}—and the standard RL baselines. Our framework demonstrates superior performance across the tested scenarios.
\end{itemize}

The remainder of this paper is structured as follows. Section.~\ref{sec.2} introduces related work in reward shaping and SSL. Section.~\ref{sec.3} presents the proposed framework alongside the necessary background on SSL. Section.~\ref{sec.3} describes the experimental setup, provides comprehensive validation results, and offers an in-depth analysis of SSL.

\section{Related Work}\label{sec.2}
Sparse reward is a common challenge in RL. In addition to reward shaping, curriculum learning and hierarchical reinforcement learning are also effective approaches to addressing this issue. Curriculum learning structures training data through a progressive sequence, transitioning the agent from elementary to increasingly intricate tasks. Recent research \cite{SayarIOK24} introduces a mechanism that generates adaptive learning objectives without domain expertise, promoting efficient learning in complex environments. Hierarchical reinforcement learning addresses the issue by decomposing a complex RL problem into subproblems, deploying specialized agents to manage distinct tiers of decision-making. \cite{BukharinLHZ25} propose a hierarchical reward modeling framework, which streamlines reward design by leveraging structural hierarchies or surrogate feedback to automatically evaluate trajectories and train reward models.

Our work seeks to leverage reward shaping, which relies on densifying the reward function and therefore mitigate challenges in exploration or credit assignment induced by sparse rewards.
There are different reward shaping methods for different objectives: Count-based methods \cite{Choi2018ContingencyAwareEI,Ostrovski2017CountBasedEW,Bellemare2016UnifyingCE} incentivize agents based on the rarity of states, while curiosity-driven exploration methods reward the agent's exploratory behavior. 
ICM \cite{Pathak2017CuriosityDrivenEB} also pursues rarity, but operationalizes it by defining curiosity as the error in the agent's ability to predict the consequences of its own actions in a visual feature space. This curiosity signal serves as an intrinsic reward, driving the agent to explore novel states. \cite{andrychowicz2017hindsight} propose Hindsight Experience Replay (HER) that enables learning from failures by treating unachieved goals as alternative goals for experience replay.
Approaches using supervised learning for reward shaping don't rely on specific objectives. The study \cite{Kumar2019RewardConditionedP} employs \textit{supervised learning} techniques, viewing non-expert trajectories collected from sub-optimal policies as optimal supervision to do the reward shaping. \cite{Memarian2021SelfSupervisedOR} leverage self-supervised methods to extract signals from trajectories while simultaneously updating the policy.

We extend supervised learning to a semi-supervised paradigm and introduce SSL techniques into the RL framework, specifically employing \textit{consistency regularization} and \textit{pseudo-labeling} \cite{Bachman2014LearningWP} in the trajectory space.
In the realm of SSL, \textit{pseudo-labeling} employs the model itself to generate artificial labels for unlabeled data, while \textit{consistency regularization} leverages unlabeled data by capitalizing on the invariance assumption—that the model should yield consistent predictions when presented with perturbed variants of identical inputs. \cite{Berthelot2019MixMatchAH,Sohn2020FixMatchSS} further improve this methodology on image classification tasks.
We will formally introduce these SSL techniques  in the later Background section and justify how our reward estimator is optimized through them.

\section{Methodology}\label{sec.3}
\begin{figure*}[h]
\begin{center}
\centerline{\includegraphics[width=0.88\textwidth]{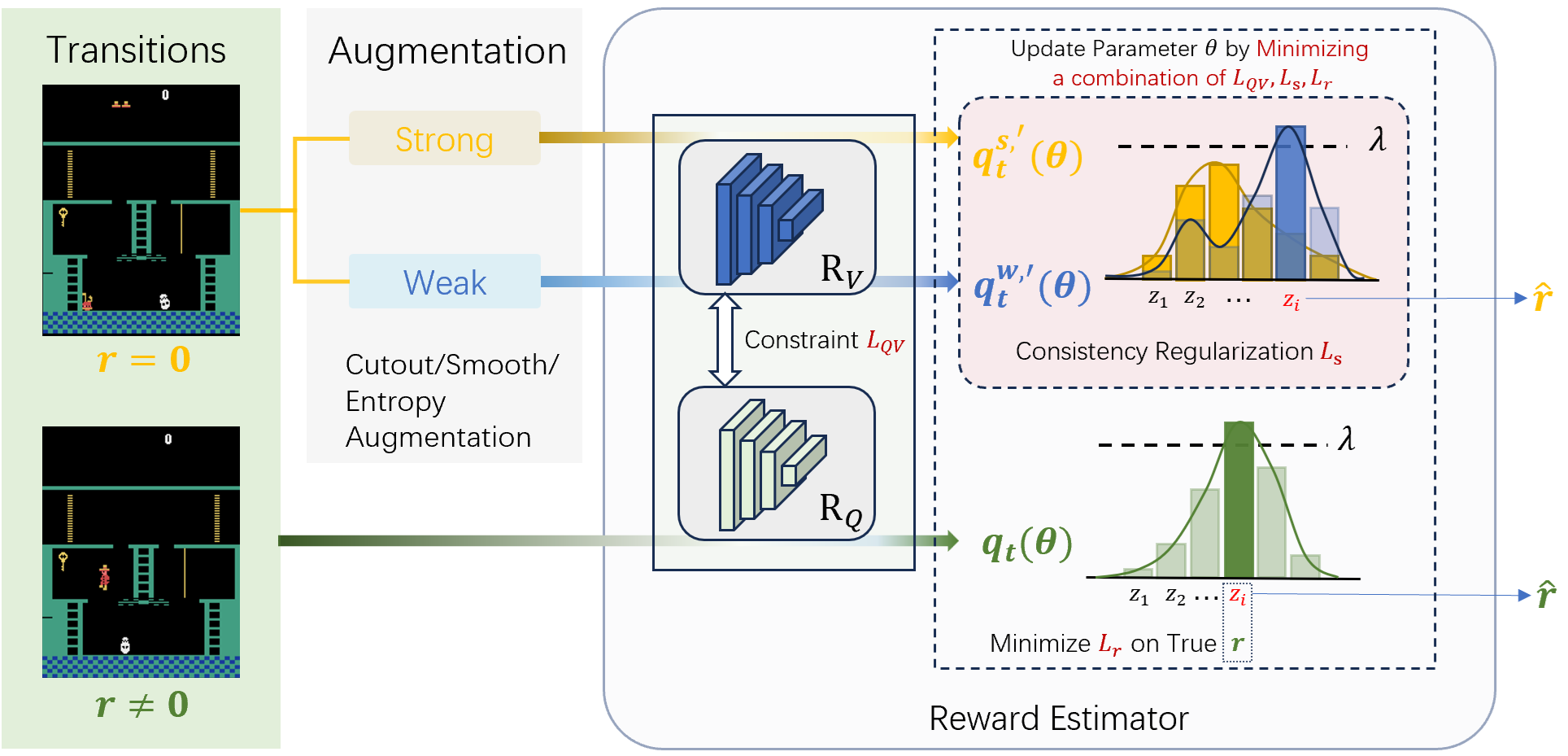}}
\caption{The figure illustrates how the SSRS method leverages both non-zero reward transitions and sparse reward transitions for reward shaping through a semi-supervised learning approach incorporating strong-weak data augmentation. The augmentation techniques follow \cite{Sohn2020FixMatchSS}, including cutout, smooth, gaussian and translation. Entropy augmentation is specified in Section.~\ref{sec.entropy}.
}
\label{fig1}
\end{center}
\end{figure*}
In this section, we introduce the proposed method and its pipeline is graphically illustrated in Figure.~\ref{fig1}, which consists of a reward estimator optimized through semi-supervised learning, and the data augmentation technique for  transitions that serves as the perturbation required for semi-supervised learning. 
We will elaborate the key parts of the method in the following subsections and the detailed steps are shown in Algorithm~\ref{alg}.

Given a Markov Decision Process (MDP) with an original reward function $r(s,a)$, reward shaping refers to modifying this reward function—either by replacing it entirely or augmenting it with an auxiliary function—to produce a new reward function $R(s,a)$. In this paper, we focus on transition-level reward shaping. Specifically, for a one-step transition at timestep $t$ represented as $\langle s_t, a_t, r_t, s_{t+1}\rangle$, the shaped transition becomes $\langle s_t, a_t, \hat{r_t}, s_{t+1}\rangle$, where $\hat{r_t}=R(s_t,a_t)$ denotes the newly assigned reward for that transition.
Since our method is of a classification nature, that is to classify zero-reward transitions by matching them to these observed non-zero rewards and accordingly assigning a corresponding reward value. Let $Z=\{z_i,\ i=1,\dots,N_z\}$ be the set of $N_z$ collected non-zero rewards, for a transition $\langle s_t, a_t, r_t, s_{t+1}\rangle$ the reward distribution $\mathbf{p}(\hat{r_t}=z|S=s_t,A=a_t)$  (where bold indicates a vector) represents the probability of assigning reward $z\in Z$ subject to the normalization condition $\sum_{z \in Z} {p}(\hat{r_t} = z | S=s_t,A=a_t) = 1$.

\subsection{Background}
\textit{Consistency regularization} and \textit{pseudo-labeling} are key components in modern state-of-the-art semi-supervised learning methods,  which have been primarily applied in classification tasks. 
To contextualize their use, we draw on the literature introduced earlier to describe these techniques. Let $\mathbf{p}(r|s,a)$  denote the predicted reward distribution generated by the model given input $s$ and $a$. \textit{Consistency regularization} relies on the assumption that the model’s output should remain stable under input perturbations, allowing the model to be optimized in a semi-supervised manner via the loss function
\begin{equation}\label{eq.consistency}
    \sum_{s,a} \left\| \mathbf{p}(r|\tilde{s},\tilde{a}) - \mathbf{p}(r|\tilde{s},\tilde{a}) \right\|_2^2,
\end{equation}
where $\tilde{s}$ and $\tilde{a}$ are stochastically perturbed versions of the original state $s$ and action $a$, and $\mathbf{p}$ denotes the model (its dependence on parameters is omitted). Because the perturbations differ, the two terms in the equation generally produce distinct values.

\textit{Pseudo-labeling} leverages the model’s own high‑confidence predictions on unlabeled data. Hard labels are obtained via $\arg\max$ operation over the predicted distribution $\mathbf{q}=\mathbf{p}(r|s,a)$—here we refer to $\mathbf{q}$ as the confidence score. Only predictions whose maximum confidence exceeds a threshold $\lambda$ are retained and optimized using the loss:
\begin{equation}
 \mathds{1}(\max(\mathbf{q}) \geq \lambda) \cdot \mathcal{H}(\mathbf{q}, \arg\max(\mathbf{q})),
\end{equation}
where $\mathds{1}(\cdot)$ is the indicator function and $\mathcal{H}$ denotes the cross-entropy.

\subsection{Semi-Supervised Reward Estimator}
The reward estimator is the very module that takes (augmented) transitions as input and outputs estimated rewards (see Figure.~\ref{fig1}). During the training phase, the estimator is optimized via consistency regularization using the cross-entropy. In the prediction phase, the final estimated reward is generated through $\arg\max$ selection, choosing the $z$ value in the reward set $Z$ corresponding to the peak of the confidence score distribution. 
Reward shaping in this study is framed as a classification task, within which the SSRS framework is designed to build a reward function that integrates a state-action-value reward distribution function $\mathbf{R}_Q(s,a)$ and a state-value reward distribution function $\mathbf{R}_V(s)$, so as to select the optimal reward from the set $Z$.
The state-value reward distribution function $\mathbf{R}_V(s): \mathcal{S} \to \Delta(Z)$ defined as $\mathbf{R}_V(s) = \mathbf{p}(\hat{r} = z | S=s)$, maps a state to a probability distribution over the reward set $Z$, where $\Delta(Z)$ is the probability simplex. Correspondingly, the state-action value reward distribution function $ \mathbf{R}_Q(s, a): \mathcal{S} \times \mathcal{A} \to \Delta(Z) $ is given by $\mathbf{R}_Q(s, a) = \mathbf{p}(\hat{r} = z | S=s, A=a)$.

Given an one-step transition $\langle s_t, a_t, r_t, s_{t+1}\rangle$, the estimated reward can be calculated from the estimations of  $\mathbf{R}_V(s)$ and $\mathbf{R}_Q(s, a)$ as follows. We first define the confidence score vector of timestep $t$ over the reward set $Z$ in Eq.(\ref{eq: 4.2.0}).
Here, $\mathbf{q}_t$ is defined similarly to $\mathbf{p}(r|s,a)$ in \textit{Consistency regularization}, although the probabilistic model here uses a combination of two reward distribution functions. Without further elaboration and due to space limitations, the performance of SSRS is quite robust to the combination coefficient, and we take 0.5 as the default.
The elements of reward set $Z$ is the true reward value collected in the trajectories. The cardinality of this set $N_z$, is a predefined hyperparameter that controls the number of reward candidates. In normal practice, $N_z$ is set slightly larger than the possible number of distinct reward signals, and we assume the additional slots (i.e., the part where the cardinality exceeds the number of distinct observed rewards) are filled with reward 0.
By the following Eq.(\ref{eq: 4.2.0}), we can get confidence vector $\mathbf{q}_t\in\mathbb{R}^{N_z}$ at timestep $t$, with each element $q_i \in \mathbf{q}_t$ corresponding to a estimated reward $z_i \in Z$, computed as
\begin{equation}\label{eq: 4.2.0}
    \mathbf{q}_t=\frac{1}{2}( \mathbf{R}_Q(s_t,a_t)+\mathbf{R}_V(s_{t+1})).
\end{equation}

The estimate reward value $z_t$ is selected with maximum confidence score above the threshold $\lambda$, i.e., $z_t=\alpha(\mathbf{q}_t,\lambda)$, where 
\begin{equation*}
    \alpha(\mathbf{q}_t, \lambda) = 
\begin{cases} 
\arg\max_{z_i}\mathbf{q}_t, &\text{for}~q_i > \lambda,\,i=1,\dots,N_z,\\
0, &\text{else}.
\end{cases}
\end{equation*}

Note that we will use $\mathbf{q}_t(\theta)$ in subsequent sections to explicitly denote its dependence on $\theta$, which consists of $\theta_1$ (the parameter of the {$\mathbf{R}_Q$ function}) and $\theta_2$ (the parameter of the {$\mathbf{R}_V$ function}).

In order to obtain the optimal $\theta$ value for accurately estimating the reward value of each transition $\langle s_t, a_t, r_t, s_{t+1} \rangle$, SSRS framework first optimizes the following loss function using only transitions with non-zero rewards, so as to minimize the discrepancy between the reward estimator's predictions and the ground truth values for these transitions:
\begin{equation}\label{eq 4.2.21}
   L_r = \frac{1}{B} \sum_{t=1 }^{ B} \mathds{1}(\max(\mathbf{q}_t(\theta))\geq \lambda)\left( {r_t} -\alpha(\mathbf{q}_t(\theta), \lambda)  \right)^2,
\end{equation}
where $B$ is the batch size of transitions.

SSRS integrates \textit{consistency regularization} and \textit{pseudo-labeling}, employing data augmentation in a manner similar to \cite{Sohn2020FixMatchSS} to optimize the reward estimator on zero-reward transitions. In sparse-reward settings, applying augmentations to both zero- and non-zero reward transitions further enhances the generalization capability of the estimator.
Explicitly, for transitions with zero rewards, the reward estimator computes the loss as shown in Eq.(\ref{eq 4.2.22}), which is the loss between the strong augmentation term and weak augmentation term following typical loss design in consistency regularization. 
Assuming continuity of trajectories in the metric space, it calculates the confidence of reward values after weak and strong augmentations, denoted by $\mathbf{q}_t^w$ and $\mathbf{q}_t^s$, respectively. The calculation of $\mathbf{q}_t^w$ and $\mathbf{q}_t^s$ follows Eq.(\ref{eq: 4.2.0}), simply replacing the input state s with its weakly or strongly augmented version. Here we omit their dependence on $\theta$ for simplicity.
For confidence values greater than the threshold in weak augmentation, the one-hot operation is performed \cite{Bachman2014LearningWP}, denoted as $\mathbf{q}_t^{w,\,\prime}$, and 
then used to compute the cross-entropy loss with the normalized confidence values greater than the threshold in strong augmentation, denoted as $\mathbf{q}_t^{s,\,\prime}$. The loss function is denoted as. 
\begin{align}
    L_s = \frac{1}{B} \sum_{t=1 }^{B} & \mathds{1}(\max(\mathbf{q}_t^s)\geq \lambda,\,\max(\mathbf{q}_t^w)\geq \lambda) \nonumber
    \\
& \cdot\mathcal{H}(\mathbf{q}_t^{w,\,\prime},\mathbf{q}_t^{s,\,\prime}), \label{eq 4.2.22}
\end{align}
where $\mathcal{H}$ denotes the cross-entropy.

Weak and strong augmentations follow \cite{Sohn2020FixMatchSS}, where we consider Gaussian noise with smaller parameters as weak augmentation, and other augmentation such as smoothing, translation, and cutout, with larger parameters as strong augmentation.
Unlike conventional image-based augmentation techniques (e.g., cutout, flipping, rotation), we exploit the unitless nature of entropy—which quantifies uncertainty in a probability distribution independently of scale or domain—to design a novel entropy augmentation method suitable for non-image tasks within SSRS. Implementation details are provided in the following section.  

\subsection{Extensions of SSRS}
Since the estimator employs two neural networks {to approximate $\mathbf{R}_V(s)$ and $\mathbf{R}_Q(s,a)$ respectively}, we introduce a constraint to regulate their outputs and stabilize the training process.
 $\mathbf{R}_V(s)$ provides a global baseline for a state, while advantage function captures the relative advantage or disadvantage of action $a$ with respect to this baseline. 
 Unlike the conventional advantage function defined over expected returns, here the "advantage" is defined solely over estimated immediate rewards.
 This separation makes it easier to assess the relative importance of specific actions, and thus compensates for the insensitivity to actions because the disturbance of data augmentation only takes place on state $s$. 
We introduce {the monotonicity} constraint in the SSRS framework as follows, which quantifies the mean square positive advantage function values. Define advantage function {$\delta_t(\theta) = \mathbf{R}_Q(s_t,a_t,\theta_1) - \mathbf{R}_V(s_{t},\theta_2)$} where $\theta$ is the combination of $\theta_1$ and $\theta_2$, we have
\begin{equation}\label{eq loss_qv}
    L_{QV} = 
\begin{cases} 
0, & \delta_t(\theta) > 0, \\
\displaystyle\frac{1}{B} \sum_{t=1 }^{ B}\left(\delta_t(\theta)\right)^2, & \text{else}.
\end{cases}
\end{equation}
Minimizing the objective loss $L_{QV}$ over the parameter $\theta$ improves SSRS performance, which will be validated in experiment section.

Upon incorporating the aforementioned extensions, the algorithm for value-based SSRS with synchronous update —\textit{i.e.}, value function update and reward shaping carry out simultaneously— is outlined in Algorithm~\ref{alg}.
{Since the reward estimator does not distinguish between augmented and non-augmented transitions, the loss function computation can be parallelized, making the algorithm implementation quite simple and straightforward.}
The agent begins by collecting experience transitions. The reward estimation network is then initialized through back-propagation, which can be performed either synchronously or asynchronously using the loss function. 
The complete SSRS loss function, as specified in Eqs.~(\ref{eq 4.2.21}), (\ref{eq loss_qv}) and (\ref{eq 4.2.22}), combines these components through the consistency coefficient $\beta$.

\subsection{Entropy Augmentation for SSRS}\label{sec.entropy}
{As a form of SSL perturbation, the implementation of entropy  augmentation primarily utilizes Shannon Entropy for computation.} \textit{Shannon Entropy} \cite{Shannon1948AMT} quantifies the amount of information required to describe or encode a random variable's possible states. In RL tasks, observations and trajectories are matrices in practice, but can also be normalized and viewed as discrete random variable. 
Consider a $m\times n$ matrix $A$ with each element $a_{ij} \geq 0,\,\forall i=1,...,m,\,j=1,...,n$, and the \textit{Shannon Entropy} of matrix $A$ is defined by
\begin{equation}\label{eq 4.1}
    H(A) := -\sum_{k=1}^{mn} p_k \log(p_k),
\end{equation}
where the {probabilities are normalized by the} matrix elements as $\displaystyle p_k =a_{ij}/\sum_i \sum_j a_{ij}$, such that $\sum_{k=1}^{mn} p_k = 1$. 

Denote state space as $\mathcal{S}\in \mathbb{R}^{m_1}$ and action space as $\mathcal{A}\in \mathbb{R}^{m_2}$, with $m_1,\,m_2 \geq 1$. Consider a trajectory $\tau$ of length {$N$ as 
$\langle s_0, a_0, r_0, s_1,\allowbreak a_1, r_1, \dots, s_N \rangle$
with $a_t \in \mathcal{A},\  s_t \in \mathcal{S}$, for all $t = 1, \dots, N$}, and rewards are given following the environment's original reward function. 
The set of all trajectories can be stacked into the matrix $\bar{\Gamma} = \begin{bmatrix} S,\,A,\,R \end{bmatrix}$ with $S\in\mathbb{R}^{N\times m_1},\,A\in\mathbb{R}^{N\times m_2}$ and $R\in\mathbb{R}^{N\times 1}$, and now we define the transform entropy augmentation $\sigma$ over matrix $\bar{\Gamma}$ as:
{\begin{equation*}
\sigma (\bar{\Gamma}|n) := \begin{bmatrix}[h^1 \cdot s^1, h^2 \cdot s^2, \dots, h^{n} \cdot s^n],\,A,\,R \end{bmatrix}, 
\end{equation*}
where $h^n \in \mathbb{R}$, equals to the entropy $H(s^n)$, matrix  $s^n$ is a submatrix of the stacked state $S$,} obtained {by equally dividing $S$ into $n$ parts along the state dimension. Here $n$ is a hyperparameter}.

\begin{algorithm}[h]
\caption{Value-based SSRS Framework} \label{alg}
\textbf{Parameter} Confidence threshold $\lambda$, reward set cardinality {$N_z$}, episodes $T$, consistency coefficient $\beta$, update probability $p_u$;
\begin{algorithmic}[1]
\STATE Initialize parameters $\theta$, $Q$ function, replay buffer $\mathcal{D}$; 
{\FOR{$t = 1$ to $T$}
    \REPEAT
        \STATE Take $a_t$ from $s_t$ using policy derived from $Q$ (e.g., $\epsilon$-greedy), observe $r_t$, $s_{t
        +1}$
        \IF{$r_t \notin \{0,r_1,\dots,r_{t-1}\}$ and $|Z|<N_z$}  
            \STATE Append $r_t$ to reward set $Z \leftarrow Z \cup \{r_t\}$
        \ENDIF
        \IF{$r_t =0$} 
            \STATE Calculate confidence $\mathbf{q}_t(\theta)$, $\mathbf{q_t^{w,\,\prime}(\theta)}$ of non-zero and zero reward transitions over set $Z$ under parameter $\theta$
            \STATE Perform reward shaping on zero-reward transitions in $\mathcal{D}$ of the $p_u$ ratio
        \ENDIF
        \STATE Choose a batch of transitions $\mathcal{B}$ from $\mathcal{D}$ and update $Q$ function with shaped reward $\hat{r}$, and set step-size $\eta$:
\begin{multline*}
    Q(s, a) \leftarrow Q(s, a) + \frac{\eta}{|\mathcal{B}|} \\
    \sum_{(s, a, \hat{r}, s') \in \mathcal{B}} \bigl( \hat{r} + \gamma \max_{a} Q(s', a) - Q(s, a) \bigr).
\end{multline*}

    \UNTIL{$s_{t+1}$ is the terminal state}
    \STATE Update parameter $\theta$ through {minimizing the} objective $L = L_{QV} +  \beta L_s + (1-\beta)L_r$
\ENDFOR
}
\end{algorithmic}
\end{algorithm}

\section{Experiment Result}\label{sec.4}
\begin{figure*}[t]
    \centering
    \subfigure[Kangaroo]{
    \includegraphics[width=0.35\textwidth]{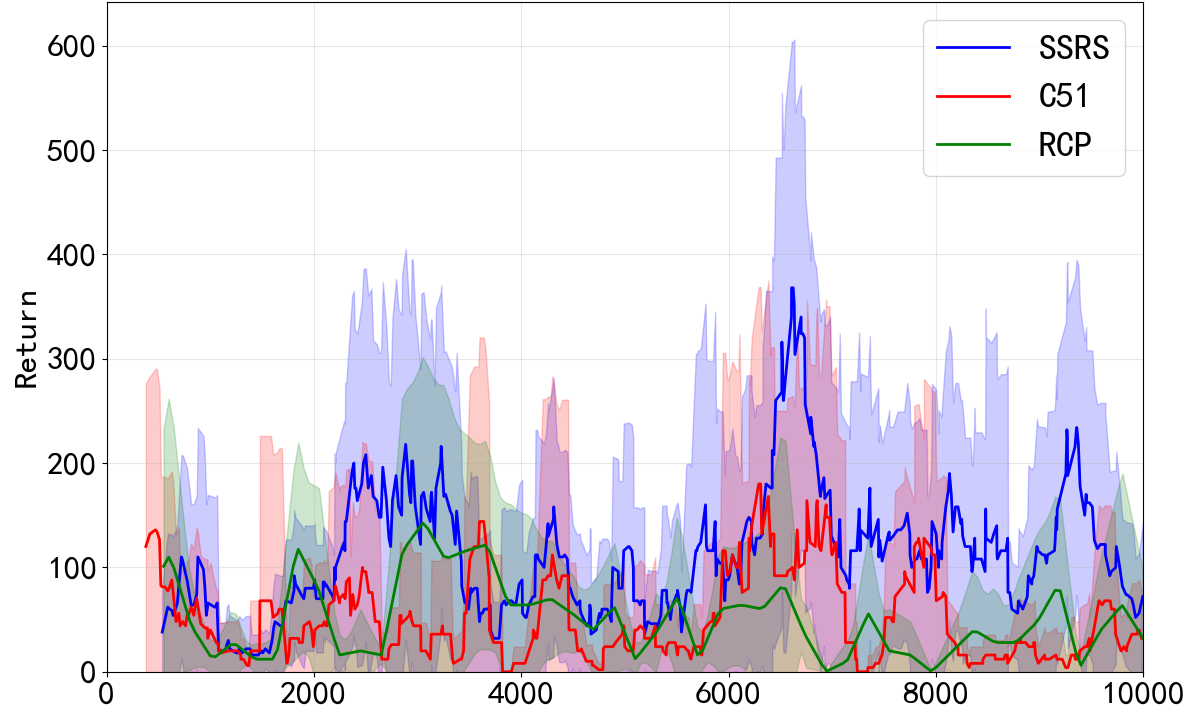}
        \label{fig.SSRS.kangaroo}
    }
    \subfigure[Seaquest]{
        \includegraphics[width=0.35\textwidth]{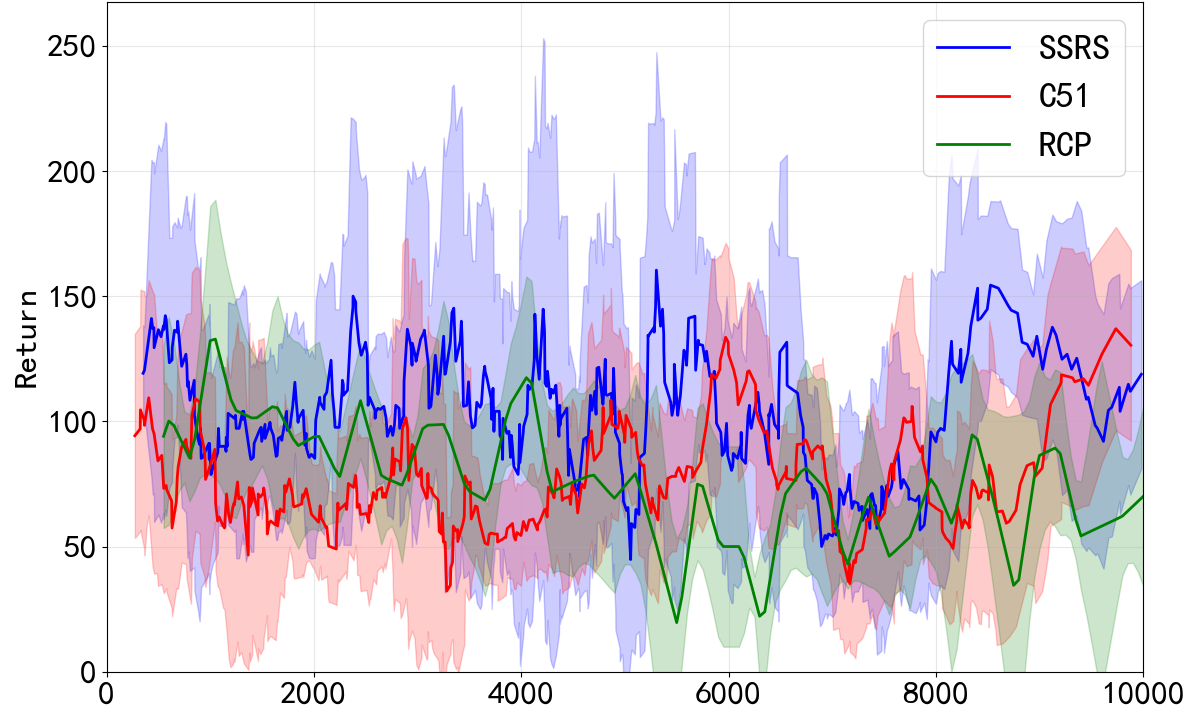}
        \label{fig.SSRS.Seaquest}
    }
    \vspace{0.1em}
    \subfigure[Montezuma's Revenge]{
        \includegraphics[width=0.30\textwidth]{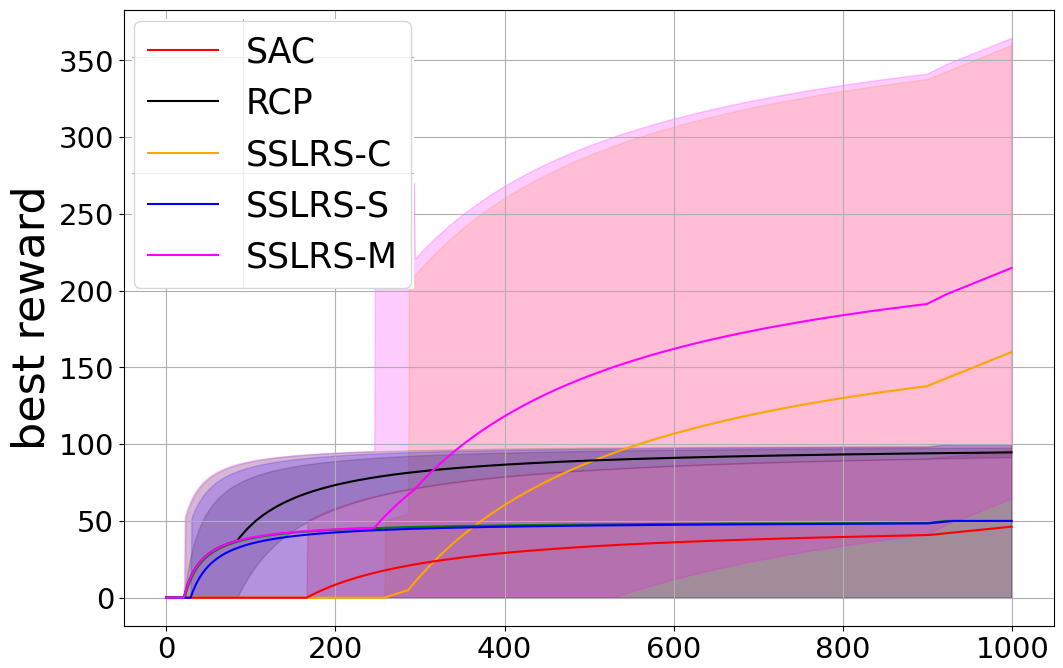}
        \label{fig.SSRS.c}
    }
    \subfigure[Venture]{
        \includegraphics[width=0.30\textwidth]{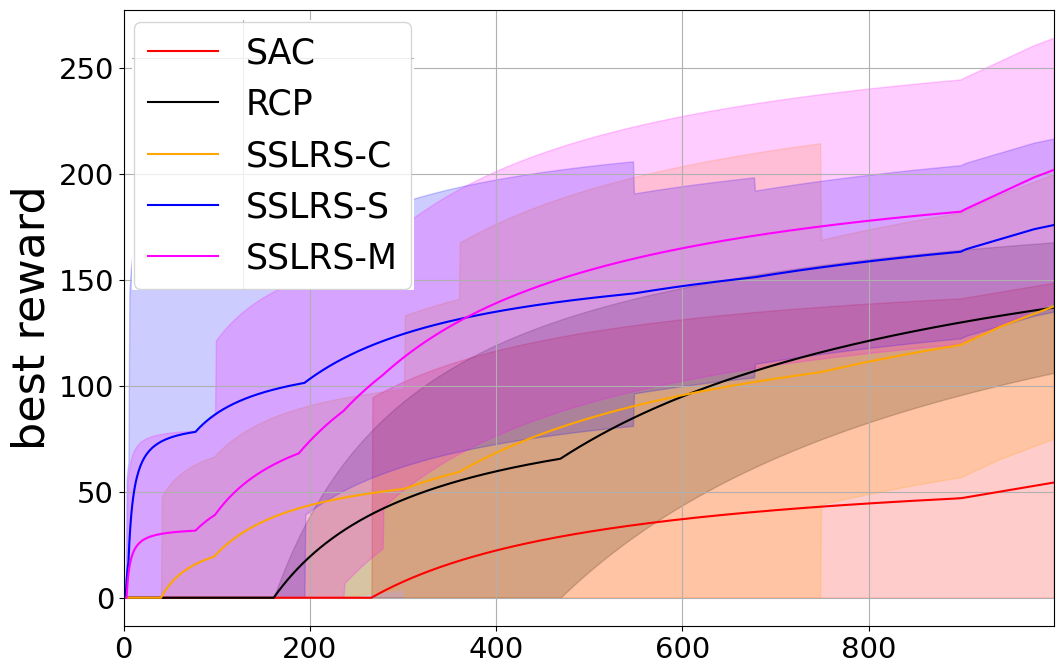}
        \label{fig.SSRS.d}
    }
    \subfigure[Hero]{
    \includegraphics[width=0.32\textwidth]{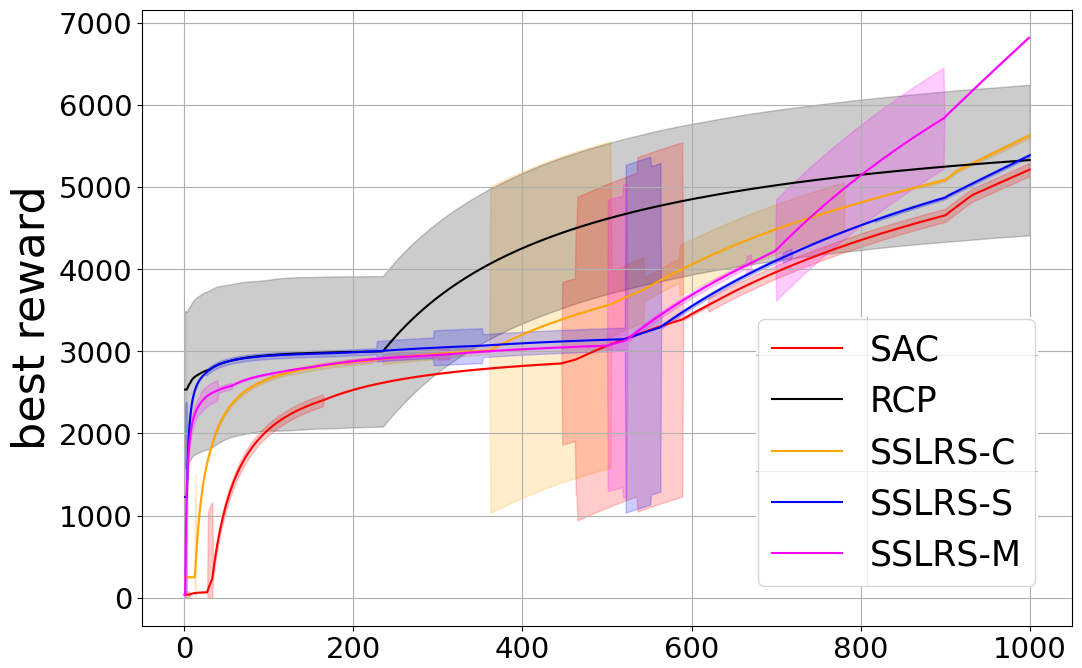}
        \label{fig.SSRS.b}
    } %
    \caption{(a)-(b) Reward curves of SSRS using C51 as the backbone, compared with the RCP algorithm.
(c)-(e) Best score curves of SSRS under different data augmentation methods (using SAC as the backbone), compared with RCP. Note: To avoid cluttered curves from multiple variants, the best score—defined as the maximum score achieved from the beginning up to the current test episode—is plotted in (c)-(e). All curves represent the mean ± standard deviation over 5 seeds. The horizontal axis denotes test steps, and training is conducted for 1000k frames.}
    \label{fig:score_curve_SSRS}
\end{figure*}
The experiments aim to validate whether the SSRS using semi-supervised learning is more efficient in reward shaping than the self-supervised method, thereby improving the best score achieved by the agent in both the Atari game environment \cite{bellemare13arcade} and the robotic manipulation environment \cite{Plappert2018MultiGoalRL}. {To demonstrate the superiority of entropy augmentation over other data augmentation methods in non-visual observation tasks, we use Random Access Memory (RAM) as the observation.} Additionally, ablation experiments are conducted to demonstrate the importance of monotonicity constraints.
We will analyze the characteristics of the trajectory space through experiments to provide insights into the underlying reasons of such improvement.

\noindent\textbf{Experimental Setup.} RAM observation refers to a representation of the Atari RL environments' internal state directly from its Random Access Memory, typically 128 bytes of RAM (\textit{i.e.}, a vector of 128 integers, each in the range [0, 255]). The implementation for this experiment is based on the Tianshou RL library \cite{tianshou}. {Across experimental seeds, a maximum of 10000 transition samples per episode are collected and appended to the buffer $\mathcal{D}$, a buffer with a capacity of 30k transitions. 
}

\subsection{Performance of SSRS}\label{section:ex_ssrs}

\begin{figure}[h]
\begin{center}
\centerline{\includegraphics[width=.7\columnwidth]{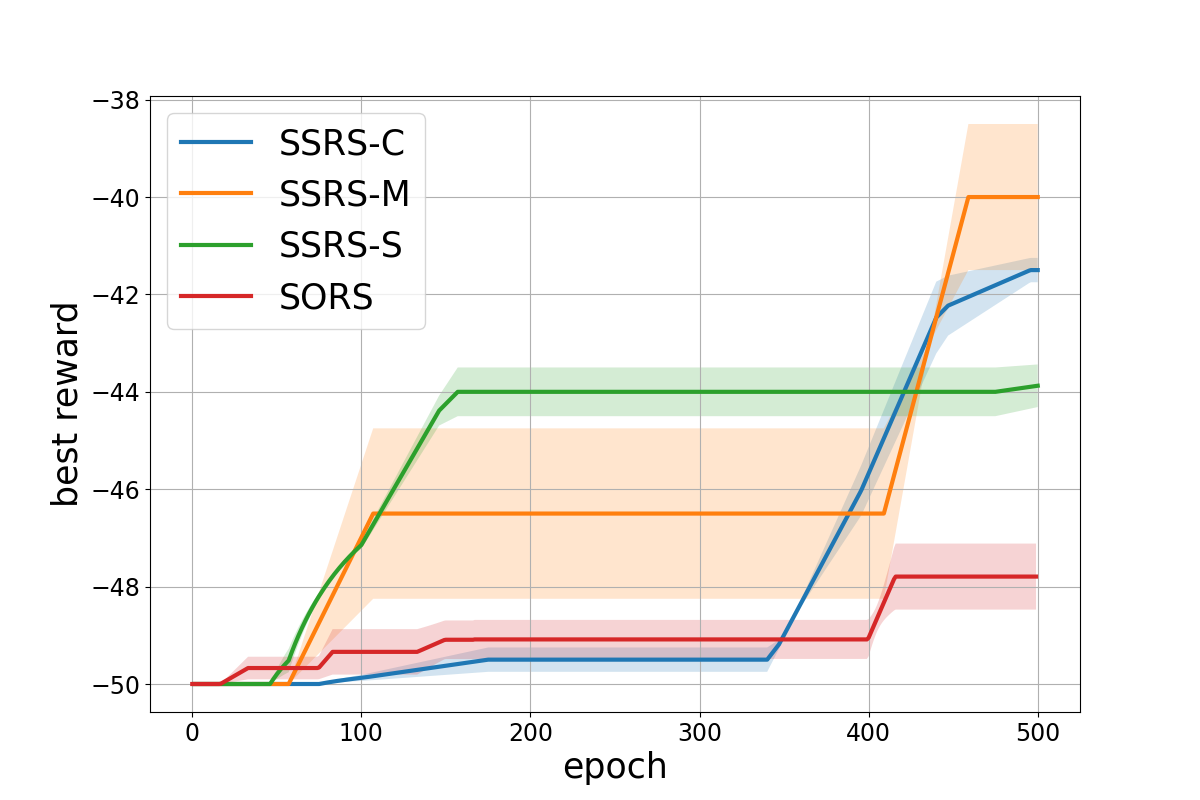}}
    \caption{This figure shows the best score curve of SSRS variants and aforementioned baseline in the robotic manipulation environment FetchReach (mean $\pm$ std over 3 seeds). The horizontal axis denotes test steps, and training is conducted for 1000k frames.
}
   \label{fig: robotic_curve}
\end{center}
\vskip -0.2in
\end{figure}

To evaluate the proposed SSRS framework, we compare it with several baseline approaches including supervised-based methods (RCP \cite{Kumar2019RewardConditionedP} and SORS \cite{Memarian2021SelfSupervisedOR}) as well as backbone reinforcement learning algorithms in both Atari and Robotics environments. 
Since this paper focuses on reward shaping, the backbone algorithms used are not the main emphasis.
The backbone algorithms for Atari and Robotics environments are Categorical DQN (C51, \cite{DBLP:conf/icml/BellemareDM17}), Soft Actor-Critic (SAC, \cite{haarnoja2019softactorcriticalgorithmsapplications})
and Deep Deterministic Policy Gradient (DDPG, \cite{Lillicrap2015ContinuousCW}).
Several variants of the SSRS framework are employed, including SSRS-S, which uses entropy augmentation, SSRS-C, which adopts the cutout method for data augmentation, and SSRS-M, which adopts the smooth method for data augmentation.

We selected five Atari games known for their sparse reward structures. For instance, in Seaquest, the player controls a submarine to rescue divers while avoiding underwater threats, with meaningful rewards occurring only upon successful rescues or eliminations. Similarly, in Kangaroo, the kangaroo character must navigate platforms to rescue its offspring, with rewards primarily tied to reaching specific objectives.  
Since these Atari environments operate on discrete action spaces, we first compare our method with the RCP algorithm, which is designed for discrete action settings. Comparisons with the SORS algorithm, developed for continuous action spaces, are deferred to later experiments in Robotics environments. The resulting learning curves and numerical results are presented in the figure and table below.

the SSRS algorithm and its variants outperform the baseline algorithm and supervised-based RCP across all five games (see Figure.~\ref{fig:score_curve_SSRS}). This is because When the proportion of transitions with non-zero rewards is low, the self-supervised method has fewer samples containing "ground truth labels" available for learning. In such cases, the SSRS approach, which leverages semi-supervised learning to infer trajectory decision boundaries, can estimate rewards from transitions more effectively than the self-supervised method. (The concept of trajectory decision boundaries will be discussed in section 'Feasibility of Semi-Supervised Learning'.) {Seaquest and Hero games are considered to be moderately reward-sparse, so we add them to the setting to evaluate the general performance of SSRS.} In terms of convergence speed, the SSRS-S variants {outperform} RCP in the Venture environment. Moreover, in games with extremely sparse rewards, such as MonteZumaRevenge, SSRS significantly {outperforms} the RCP algorithm, achieving nearly 2 times the best score in MonteZumaRevenge (see Figure.~\ref{fig.SSRS.c}). 
In the FetchReach environment in Robotics{, where the reward signal is binary, \textit{i.e.}, the reward is -1 if the end effector hasn’t reached its final target position, and 0 if the end effector is in the final target position,} the SSRS framework outperforms the self-supervised method SORS (see Figure.~\ref{fig: robotic_curve}). 

We also examine the performance of SSRS when using different data augmentation methods as perturbations for SSL, aiming to investigate whether the proposed entropy augmentation provides greater improvement in terms of final best score for SSRS compared to other traditional data augmentation methods (e.g., flip, scale, translation), which are not shown here due to space limitations.

Additionally, we conduct a series of experiments on the SSRS framework under conditions with and without monotonicity constraint.
The average best score obtained is presented in Table.~\ref{tab: L2 score}. In three environments, SSRS with a monotonicity constraint {achieves} higher best scores compared to SSRS without a monotonicity constraint.

\begin{table}[t]
\caption{Average best score of SSRS framework with monotonicity (SSRS-ST) 
constraint and without monotonicity constraint (SSRS-NST) in 4 environments (mean $\pm$ std over 3 seeds).}
\label{tab: L2 score}
\begin{center}
\begin{small}
\begin{sc}
\begin{tabular}{l|ll}
\hline
avg best score & \textbf{SSRS-ST} & \textbf{SSLRL-NST} \\ \hline
Seaquest       & \textbf{476.6} $\pm$ 85      & 447.5  $\pm$ 18.9             \\
Hero           & \textbf{8825}   $\pm$ 2317   & 7596.6  $\pm$ 81.4             \\
Montezuma      & \textbf{166.6}  $\pm$ 104   & 100 $\pm$ 70.7                \\
Venture        & 200  $\pm$ 0              & \textbf{262.5}  $\pm$ 47.9      \\ \hline
\end{tabular}
\end{sc}
\end{small}
\end{center}
\end{table}

\subsection{Feasibility of Semi-Supervised Learning}\label{subsection 5.4}

\begin{figure}[ht]
    \begin{center}    \centerline{\includegraphics[width=.7\columnwidth]{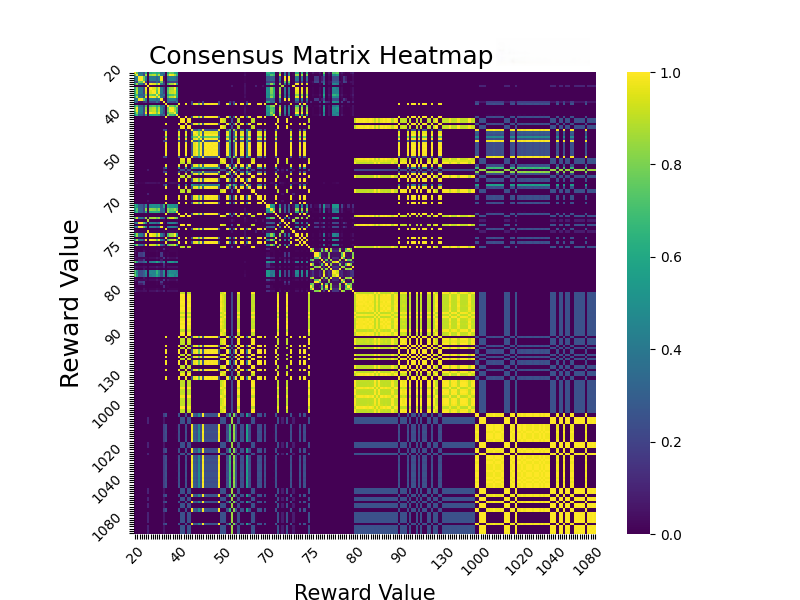}}
    \caption{This figure shows the consensus matrix of reward distribution in Hero after 1000 episodes iterations, which indicates that the distribution of transitions along the reward dimension in the trajectory space exhibits a certain clustering property. The cluster method used is Gaussian Mixture Models, and the number of iterations for consensus matrix is 100.
}
    \label{fig: consensus}
    \end{center}
\end{figure} 

To further investigate why SSL method can infer rewards from transitions more effectively than self-supervised methods—and why SSL techniques can be applied to reward shaping, we conduct the following experiment and analysis.
We examine the transition distribution and obtain the consensus matrix $\mathcal{C}$ in Figure.~\ref{fig: consensus}. Each element $ a_{ij} $ $\in\mathcal{C},0\leq a_{ij}\leq 1,$ is frequency at which {the two transitions} $\tau_i $ and $\tau_j $ are assigned to the same shaped reward value. Note that we use estimated reward values for labels. 

Larger estimated reward values, i.e., rewards given for achieving long-term goals, exhibit clearer diagonal boundaries compared to smaller reward values. 
However, the decision boundaries inside high reward values are more ambiguous, with only one clear diagonal block, aligning with the clustering assumption of SSL. 
SSL methods relies on the smoothness assumption and clustering assumption \cite{Ouali2020AnOO} on the augmented data, in this case, the trajectories.
To be specific, for classification models, the smoothness assumption indicates if two points $x_1, x_2$ reside in a high-density region are close, then
 so should be their corresponding outputs $y_1, y_2$. And clustering assumption indicates that samples from the same class are closer to each other, while there are significant boundaries between samples from different classes and the decision boundary of the model should be as far as possible from regions of high data density. 
Notably, this observed ambiguity in high reward values has minimal practical impact on SSRS, as the reward estimator outputs these high rewards with negligible confidence probabilities.

Unlike conventional data augmentation techniques that may introduce excessive perturbations to non-image data—violating the smoothness and clustering assumptions—  entropy augmentation better preserves these assumptions in trajectory space. This is crucial to the step of calculating estimated reward, where the $\mathbf{R}_V$ network and $\mathbf{R}_Q$ network are not able to separate transitions to learn a good representation of the space. From the perspective of the teacher-student model \cite{10.5555/3294771.3294885}, such a `muddled' teacher would lead the student, \textit{i.e.} policy module, to learn an unstable policy.

From Figure.~\ref{fig: consensus}, it can be observed that the three `phases' of significance in the consensus matrix plot correspond to the downward trend of the reward distribution probability. This can be understood from the perspective that as the estimated reward value increases, the {randomness} of the transition decreases, leading to greater clustering significance in the trajectory space, and \textit{vice versa}.

{
\subsection{Hyperparameter Robustness Study}
Regarding the value of the cardinality $N_z$ of the reward candidate set $Z$, it is set to be slightly larger than the possible number of distinct reward signals after preliminary experiments, which avoids both overcomputation and insufficient reward exploration.

SSRS introduces three hyperparameters, the update probability $p_u$, confidence threshold $\lambda$, and loss coefficient $\beta$. 
For supervised-based reward shaping \cite{Memarian2021SelfSupervisedOR,Kumar2019RewardConditionedP,Peng2019AdvantageWeightedRS}, the parameter controlling the update portion is natural, and experimental results (see Figure.~\ref{fig: hyperparameter}) also demonstrate its limited impact on the robustness of this type of model, except when $p_u$ approaches zero. In this case, no transitions undergo reward shaping, thus making SSRS unattainable to improve performance.
The confidence threshold $\lambda$ is one of the few hyperparameters in pseudo-labeling models \cite{Lee2013PseudoLabelT,Sohn2020FixMatchSS}, and we also investigate its influence under the reinforcement learning framework. Similar to that observed in computer vision tasks, performance peaks occur when the confidence threshold value is between 0.8-0.9.
Finally, the loss coefficient $\beta$ has been shown (see Table.~\ref{tab:consistency_coefficient}) to primarily affect the training process, with minimal impact on the final results. 

\begin{figure}[ht]
    \begin{center}    \centerline{\includegraphics[width=.7\columnwidth]{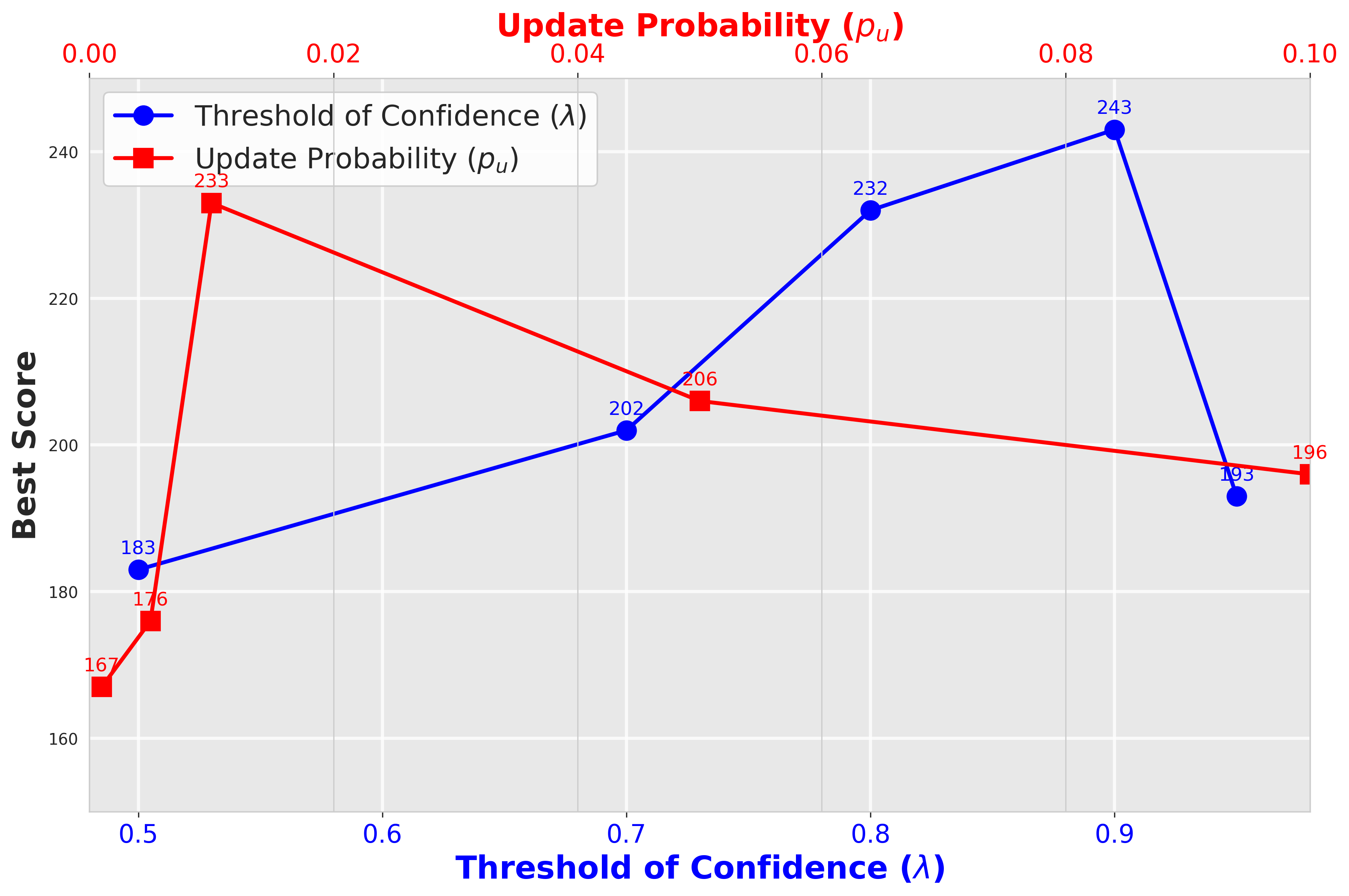}}
    \caption{This figure illustrates the impact of the hyperparameters update probability $p_u$ and confidence threshold $\lambda$ on SSRS (using the Venture environment as an example). The lower x-axis represents the values of the confidence threshold, while the upper x-axis represents the values of the update probability.
}
    \label{fig: hyperparameter}
    \end{center}
\end{figure} 
\begin{table}[htbp]
\centering
\caption{Performance of SSRS with different consistency coefficients $\beta$ in the Venture environment. This table shows the best scores achieved and the average corresponding training epochs when varying the values of $\beta$.}
\label{tab:consistency_coefficient}
\begin{tabular}{c|ccccc}
\hline
\textbf{Consistency Coefficient $\beta$} & 0.3 & 0.4 & 0.5 & 0.6 & 0.7 \\ \hline
Best score                       & 226         & 230         & 223         & 233         & 236         \\ 
In epoch                         & 854         & 853         & 862         & 877         & 892         \\ \hline
\end{tabular}
\end{table}
}

\section{Conclusion}

In this paper, we propose a Semi-Supervised Reward Shaping (SSRS) framework which utilizes zero reward trajectories by employing SSL technique. Additionally, we propose the entropy augmentation method for consistency regularization and a monotonicity constraint over modules of reward estimator. Our model outperforms previous supervised-based RCP and SORS methods in Atari environments, with a maximum of twice increase in reaching higher best scores. Moreover, the entropy augmentation enhances performance showcasing a 15.8\% increase in best score compared to other augmentation methods.

Moreover, it's worth noting that SSRS also introduces several hyperparameters, {such as the update probability of trajectories in the experience replay buffer $p_u$. Although the tuning of these hyperparameters has been studied in the literature of SSL, these hyperparameters can still provide valuable insights for the field of reinforcement learning.} In this class of reward shaping algorithms, the tuning of $p_u$ reflects the balance between exploration and exploitation in that, a higher $p_u$ encourages the agent to exploit knowledge learned from the reward function, while a lower $p_u$  prompts the agent to explore according to its original policy. Similar issues were also highlighted in \cite{Peng2019AdvantageWeightedRS,Kumar2019RewardConditionedP}, where the problem of controlling the reward shaping ratio arises. Like our paper, the balance shifts towards deciding whether the agent should exploit the reward from the estimator or prioritize exploration. We believe that a theoretical analysis of the dynamic relationship between the optimality of the estimator and the exploration-exploitation ratio would be a valuable direction for future work.

\bibliographystyle{IEEEtran}
\bibliography{references}

@inproceedings{Bachman2014LearningWP,
  author    = {Bachman, Philip and Alsharif, Ouais and Precup, Doina},
  title     = {Learning with Pseudo-Ensembles},
  booktitle = {Proceedings of the 28th International Conference on Neural Information Processing Systems},
publisher    = {OpenReview.net},
  series    = {NIPS '14},
  year      = {2014},
  volume    = {2},
  pages     = {3365--3373},
  address   = {Cambridge, MA, USA},
  location  = {Montreal, Canada}
}

@article{Kumar2019RewardConditionedP,
  title={Reward-Conditioned Policies},
  author={Aviral Kumar and Xue Bin Peng and Sergey Levine},
  journal={ArXiv},
  year={2019},
  volume={abs/1912.13465},
  url={https://arxiv.org/abs/1912.13465},
pages={-},
numpages={-}
}

@inproceedings{Ng1999PolicyIU,
author = {Ng, Andrew Y. and Harada, Daishi and Russell, Stuart J.},
title = {Policy Invariance Under Reward Transformations: Theory and Application to Reward Shaping},
year = {1999},
isbn = {1558606122},
publisher = {Morgan Kaufmann Publishers Inc.},
address = {San Francisco, CA, USA},
booktitle = {Proceedings of the Sixteenth International Conference on Machine Learning},
pages = {278–287},
numpages = {10},
series = {ICML '99}
}

@inproceedings{Choi2018ContingencyAwareEI,
  author       = {Jongwook Choi and
                  Yijie Guo and
                  Marcin Moczulski and
                  Junhyuk Oh and
                  Neal Wu and
                  Mohammad Norouzi and
                  Honglak Lee},
  title        = {Contingency-Aware Exploration in Reinforcement Learning},
  booktitle    = {7th International Conference on Learning Representations, {ICLR} 2019,
                  New Orleans, LA, USA, May 6-9, 2019},
  publisher    = {OpenReview.net},
  year         = {2019},
  url          = {https://openreview.net/forum?id=HyxGB2AcY7},
  bibsource    = {dblp computer science bibliography, https://dblp.org},
address={-},
pages={-},
numpages={-}
}

@inproceedings{Ostrovski2017CountBasedEW,
author = {Ostrovski, Georg and Bellemare, Marc G. and van den Oord, A\"{a}ron and Munos, R\'{e}mi},
title = {Count-based exploration with neural density models},
year = {2017},
publisher = {JMLR.org},
booktitle = {Proceedings of the 34th International Conference on Machine Learning - Volume 70},
pages = {2721–2730},
numpages = {10},
location = {Sydney, NSW, Australia},
series = {ICML'17},
address={Sydney, NSW, Australia}
}

@inproceedings{Bellemare2016UnifyingCE,
  author    = {Marc G. Bellemare and Sriram Srinivasan and Georg Ostrovski and Tom Schaul and David Saxton and R{\'e}mi Munos},
  title     = {Unifying Count-Based Exploration and Intrinsic Motivation},
  booktitle = {Proceedings of the 30th International Conference on Neural Information Processing Systems},
publisher    = {OpenReview.net},
address={-},
  series    = {NIPS '16},
  year      = {2016},
  pages     = {1471--1479}
}

@InProceedings{Pathak2017CuriosityDrivenEB,
  title = 	 {Curiosity-driven Exploration by Self-supervised Prediction},
  author =       {Deepak Pathak and Pulkit Agrawal and Alexei A. Efros and Trevor Darrell},
  booktitle = 	 {Proceedings of the 34th International Conference on Machine Learning},
  pages = 	 {2778--2787},
  year = 	 {2017},
  editor = 	 {Precup, Doina and Teh, Yee Whye},
  volume = 	 {70},
  series = 	 {Proceedings of Machine Learning Research},
  month = 	 {06--11 Aug},
  publisher =    {PMLR},
  url = 	 {https://proceedings.mlr.press/v70/pathak17a.html},
address={Sydney, Australia}
}

@article{Memarian2021SelfSupervisedOR,
  title={Self-Supervised Online Reward Shaping in Sparse-Reward Environments},
  author={Farzaneh Memarian and Wonjoon Goo and Rudolf Lioutikov and Ufuk Topcu and Scott Niekum},
  journal={2021 IEEE/RSJ International Conference on Intelligent Robots and Systems (IROS)},
  year={2021},
  pages={2369-2375},
volume={-},
number={-}
}

@article{Ouali2020AnOO,
author = {Yang, Xiangli and Song, Zixing and King, Irwin and Xu, Zenglin},
title = {A Survey on Deep Semi-Supervised Learning},
year = {2023},
issue_date = {Sept. 2023},
publisher = {IEEE Educational Activities Department},
address = {USA},
volume = {35},
number = {9},
issn = {1041-4347},
url = {https://doi.org/10.1109/TKDE.2022.3220219},
doi = {10.1109/TKDE.2022.3220219},
journal = {IEEE Trans. on Knowl. and Data Eng.},
month = sep,
pages = {8934–8954},
numpages = {21}
}

@article{tianshou,
  author  = {Jiayi Weng and Huayu Chen and Dong Yan and Kaichao You and Alexis Duburcq and Minghao Zhang and Yi Su and Hang Su and Jun Zhu},
  title   = {Tianshou: A Highly Modularized Deep Reinforcement Learning Library},
  journal = {Journal of Machine Learning Research},
  year    = {2022},
  volume  = {23},
  number  = {267},
  pages   = {1--6},
  url     = {http://jmlr.org/papers/v23/21-1127.html}
}

@article{Shannon1948AMT,
  title={A mathematical theory of communication},
  author={Claude E. Shannon},
  journal={Bell Syst. Tech. J.},
  year={1948},
  volume={27},
  pages={623-656},

  number={3},
  doi={10.1002/j.1538-7305.1948.tb01338.x}}

@article{Plappert2018MultiGoalRL,
  title={Multi-Goal Reinforcement Learning: Challenging Robotics Environments and Request for Research},
  author={Matthias Plappert and Marcin Andrychowicz and Alex Ray and Bob McGrew and Bowen Baker and Glenn Powell and Jonas Schneider and Joshua Tobin and Maciek Chociej and Peter Welinder and Vikash Kumar and Wojciech Zaremba},
  journal={ArXiv},
  year={2018},
  volume={abs/1802.09464},
  url={https://arxiv.org/abs/1802.09464},
pages={-},
numpages={-}
}

@article{Peng2019AdvantageWeightedRS,
  title={Advantage-Weighted Regression: Simple and Scalable Off-Policy Reinforcement Learning},
  author={Xue Bin Peng and Aviral Kumar and Grace Zhang and Sergey Levine},
  journal={ArXiv},
  year={2019},
  volume={abs/1910.00177},
  url={https://arxiv.org/abs/1910.00177},
pages={-},
numpages={-}
}

@article{andrychowicz2017hindsight,
  title={Hindsight experience replay},
  author={Andrychowicz, Marcin and Wolski, Filip and Ray, Alex and Schneider, Jonas and Fong, Rachel and Welinder, Peter and McGrew, Bob and Tobin, Josh and Pieter Abbeel, OpenAI and Zaremba, Wojciech},
  journal={Advances in Neural Information Processing Systems},
  volume={30},
pages={-},
  year={2017}
}

@Article{bellemare13arcade,
    author = {{Bellemare}, M.~G. and {Naddaf}, Y. and {Veness}, J. and {Bowling}, M.},
    title = {The Arcade Learning Environment: An Evaluation Platform for General Agents},
    journal = {Journal of Artificial Intelligence Research},
    year = "2013",
    volume = "47",
    pages = "253--279",
}

@article{Kober2013ReinforcementLI,
  title={Reinforcement learning in robotics: A survey},
  author={Kober, Jens and Bagnell, J Andrew and Peters, Jan},
  journal={The International Journal of Robotics Research},
  volume={32},
  number={11},
  pages={1238--1274},
  year={2013},
  publisher={SAGE Publications Sage UK: London, England}
}

@article{Lillicrap2015ContinuousCW,
  title={Continuous control with deep reinforcement learning},
  author={Lillicrap, Timothy P and Hunt, Jonathan J and Pritzel, Alexander and Heess, Nicolas and Erez, Tom and Tassa, Yuval and Silver, David and Wierstra, Daan},
  journal={arXiv preprint arXiv:1509.02971},
  year={2015},
number={-},
volume={-},
pages={-},
numpages={-}
}

@inproceedings{Sohn2020FixMatchSS,
author = {Sohn, Kihyuk and Berthelot, David and Li, Chun-Liang and Zhang, Zizhao and Carlini, Nicholas and Cubuk, Ekin D. and Kurakin, Alex and Zhang, Han and Raffel, Colin},
title = {FixMatch: simplifying semi-supervised learning with consistency and confidence},
year = {2020},
isbn = {9781713829546},
publisher = {Curran Associates Inc.},
address = {Red Hook, NY, USA},
booktitle = {Proceedings of the 34th International Conference on Neural Information Processing Systems},
articleno = {51},
numpages = {13},
location = {Vancouver, BC, Canada},
series = {NIPS '20}
}

@inbook{Berthelot2019MixMatchAH,
author = {Berthelot, David and Carlini, Nicholas and Goodfellow, Ian and Oliver, Avital and Papernot, Nicolas and Raffel, Colin},
title = {MixMatch: a holistic approach to semi-supervised learning},
year = {2019},
publisher = {Curran Associates Inc.},
address = {Red Hook, NY, USA},
booktitle = {Proceedings of the 33rd International Conference on Neural Information Processing Systems},
articleno = {454},
chapter={-},
pages={-},
numpages = {11}

}

@article{Lee2013PseudoLabelT,
author = {Lee, Dong-Hyun},
year = {2013},
month = {07},
pages = {-},
title = {Pseudo-Label : The Simple and Efficient Semi-Supervised Learning Method for Deep Neural Networks},
journal = {ICML 2013 Workshop : Challenges in Representation Learning (WREPL)},
volume={-}
}

@misc{haarnoja2019softactorcriticalgorithmsapplications,
      title={Soft Actor-Critic Algorithms and Applications}, 
      author={Tuomas Haarnoja and Aurick Zhou and Kristian Hartikainen and George Tucker and Sehoon Ha and Jie Tan and Vikash Kumar and Henry Zhu and Abhishek Gupta and Pieter Abbeel and Sergey Levine},
      year={2019},
      eprint={1812.05905},
      archivePrefix={arXiv},
      primaryClass={cs.LG},
      url={https://arxiv.org/abs/1812.05905}, 
}

@inproceedings{10.5555/3294771.3294885,
author = {Tarvainen, Antti and Valpola, Harri},
title = {Mean teachers are better role models: Weight-averaged consistency targets improve semi-supervised deep learning results},
year = {2017},
isbn = {9781510860964},
publisher = {Curran Associates Inc.},
address = {Red Hook, NY, USA},
booktitle = {Proceedings of the 31st International Conference on Neural Information Processing Systems},
pages = {1195–1204},
numpages = {10},
location = {Long Beach, California, USA},
series = {NIPS'17}
}

@inproceedings{DBLP:conf/icml/BellemareDM17,
  author       = {Marc G. Bellemare and
                  Will Dabney and
                  R{\'{e}}mi Munos},
  editor       = {Doina Precup and
                  Yee Whye Teh},
  title        = {A Distributional Perspective on Reinforcement Learning},
  booktitle    = {Proceedings of the 34th International Conference on Machine Learning,
                  {ICML} 2017, Sydney, NSW, Australia, 6-11 August 2017},
  series       = {Proceedings of Machine Learning Research},
  volume       = {70},
  pages        = {449--458},
  publisher    = {{PMLR}},
  year         = {2017},
}

@inproceedings{NIPS2017_32fdab65,
 author = {Hadfield-Menell, Dylan and Milli, Smitha and Abbeel, Pieter and Russell, Stuart J and Dragan, Anca},
 booktitle = {Advances in Neural Information Processing Systems},
 editor = {I. Guyon and U. Von Luxburg and S. Bengio and H. Wallach and R. Fergus and S. Vishwanathan and R. Garnett},
 pages = {},
 publisher = {Curran Associates, Inc.},
 title = {Inverse Reward Design},
 volume = {30},
 year = {2017}
}

@inproceedings{SayarIOK24,
  author       = {Erdi Sayar and
                  Giovanni Iacca and
                  Ozgur S. Oguz and
                  Alois Knoll},
  editor       = {Amir Globersons and
                  Lester Mackey and
                  Danielle Belgrave and
                  Angela Fan and
                  Ulrich Paquet and
                  Jakub M. Tomczak and
                  Cheng Zhang},
  title        = {Diffusion-based Curriculum Reinforcement Learning},
  booktitle    = {Advances in Neural Information Processing Systems 38: Annual Conference
                  on Neural Information Processing Systems 2024, NeurIPS 2024, Vancouver,
                  BC, Canada, December 10 - 15, 2024},
  year         = {2024},
}

@inproceedings{BukharinLHZ25,
  author       = {Alexander Bukharin and
                  Yixiao Li and
                  Pengcheng He and
                  Tuo Zhao},
  title        = {Deep Reinforcement Learning from Hierarchical Preference Design},
  booktitle    = {Forty-second International Conference on Machine Learning, {ICML}
                  2025, Vancouver, BC, Canada, July 13-19, 2025},
  publisher    = {OpenReview.net},
  year         = {2025},

}
\end{document}